\documentclass{bmvc2k}

\usepackage{graphicx}
\graphicspath{{images/}}
\usepackage{amsmath,amssymb} 
\usepackage{times}
\usepackage{epsfig}
\usepackage{xspace}
\usepackage{multirow}
\usepackage{url}

\usepackage[section]{placeins}

\usepackage{ulem}

\usepackage{array,booktabs}
\newcolumntype{C}[1]{>{\centering\let\newline\\\arraybackslash\hspace{0pt}}p{#1}}
\newcolumntype{L}[1]{>{\let\newline\\\arraybackslash\hspace{0pt}}p{#1}}

\definecolor{darkblue}{RGB}{128, 0, 0}
\newif\ifhighlightchanges
\newcommand{\changemarker}[1]{%
  \ifhighlightchanges
    \textcolor{darkblue}{#1}%
  \else
    #1%
  \fi
}

\newcommand{\R}{\mathbb R}

\newcommand{\methname}{POMP\xspace}

\makeatletter
\DeclareRobustCommand\onedot{\futurelet\@let@token\@onedot}
\def\@onedot{\ifx\@let@token.\else.\null\fi\xspace}

\def\eg{\emph{e.g}\onedot} 
\def\ie{\emph{i.e}\onedot}

\def\etal{\emph{et al}\onedot}
\makeatother

\title{POMP: Pomcp-based Online Motion Planning for active visual search in indoor environments}

\addauthor{Yiming Wang}{Yiming.Wang@iit.it}{1}
\addauthor{Francesco Giuliari}{francesco.giuliari@univr.it}{3}
\addauthor{Riccardo Berra}{riccardo.berra@studenti.univr.it}{3}
\addauthor{Alberto Castellini}{alberto.castellini@univr.it}{3}
\addauthor{Alessio Del Bue}{alessio.delbue@iit.it}{2}
\addauthor{Alessandro Farinelli}{alessandro.farinelli@univr.it}{3}
\addauthor{Marco Cristani}{marco.cristani@univr.it}{3}
\addauthor{Francesco Setti}{francesco.setti@univr.it}{3}

\addinstitution{
    Visual Geometry and Modelling (VGM) \\
    Istituto Italiano di Tecnologia (IIT)\\
    Genova, Italy
}
\addinstitution{
    Pattern Analysis and Computer Vision (PAVIS) \\
    Istituto Italiano di Tecnologia (IIT)\\
    Genova, Italy
}
\addinstitution{
    Department of Computer Science\\
    University of Verona\\
    Verona, Italy
}

\runninghead{Wang \etal}{POMP: Pomcp-based Online Motion Planning for AVS}
\begin{document}
\normalem


\maketitle

\begin{abstract}
In this paper we focus on the problem of learning an optimal policy for Active Visual Search (AVS) of objects in known indoor environments with an online setup.
Our \methname method uses as input the current pose of an agent (\eg a robot) and a RGB-D frame. The task is to plan the next move that brings the agent closer to the target object. We model this problem as a Partially Observable Markov Decision Process solved by a Monte-Carlo planning approach.
This allows us to make decisions on the next moves by iterating over the known scenario at hand, exploring the environment and searching for the object at the same time.
Differently from the current state of the art in Reinforcement Learning, \methname does not require extensive and expensive (in time and computation) labelled data so being very agile in solving AVS in small and medium real scenarios.
We only require the information of the floormap of the environment, an information usually available or that can be easily extracted from an a priori single exploration run.
We validate our method on the publicly available AVD benchmark, achieving an average success rate of 0.76 with an average path length of 17.1, performing close to the state of the art but without any training needed. Additionally, we show experimentally the robustness of our method when the quality of the object detection goes from ideal to faulty.
\end{abstract}

\FloatBarrier
\section{Introduction}

Autonomous navigation in outdoor and urban environments has received major attention in the vision and robotic communities, mostly driven by investments from the automotive industry.
\changemarker{Differently, less attention has been dedicated to indoor navigation where the diversity in the environment structure is providing open and new scientific challenges.}
%

\begin{figure}[!t]
    \centering
    \includegraphics[width=0.9\linewidth]{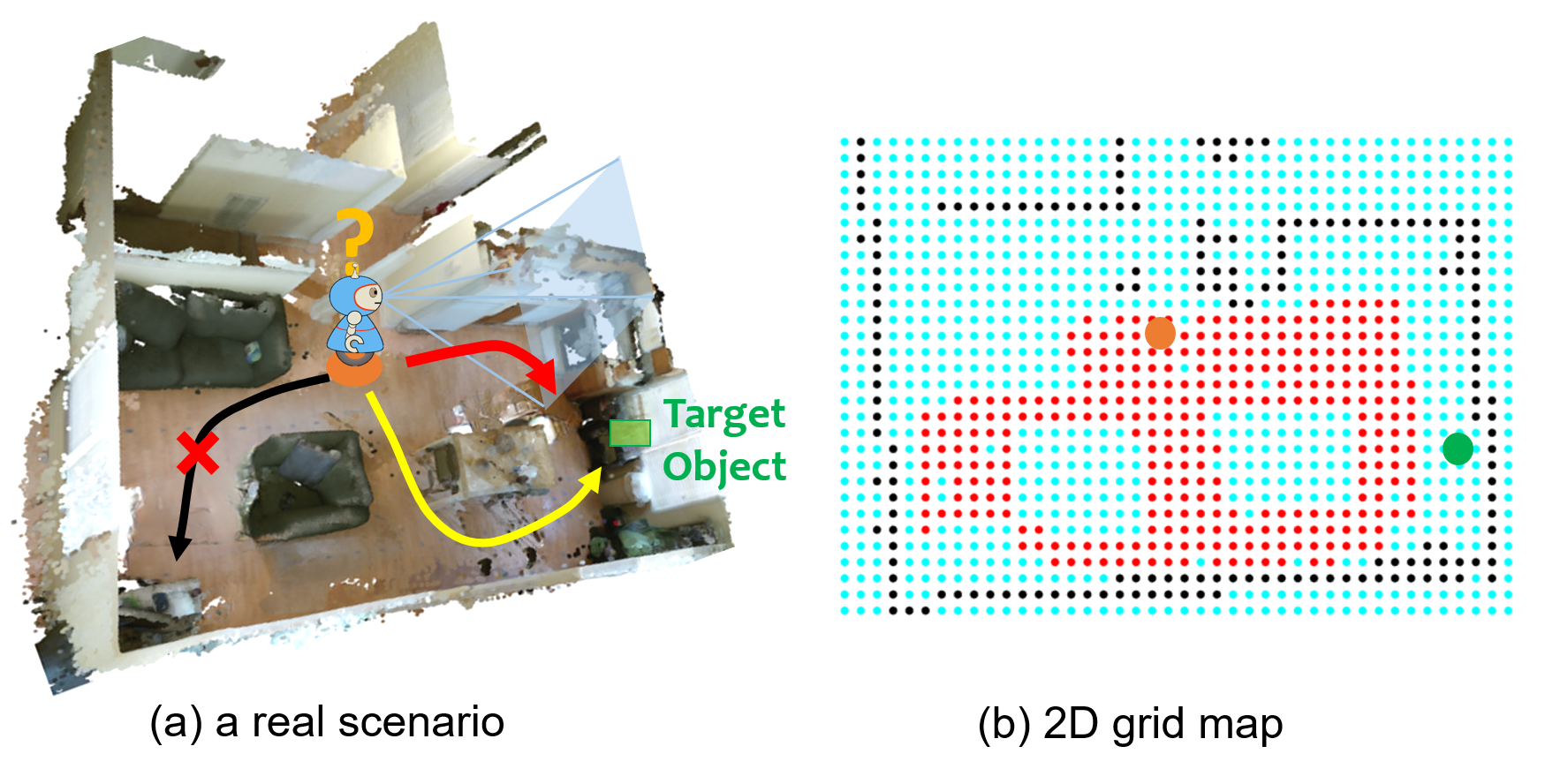}
    \vspace{-0.3cm}
    \caption{An agent is initialised (highlighted in orange) in a known environment with the task of visually searching for a target object (highlighted in green), \ie to have the object detected on the captured image. (a) The agent navigates in a real scenario, driven by a motion policy to detect the target object with the shortest travelled path (highlighted in red), to avoid longer trajectories (in yellow) or missing entirely the target (in black). (b) The 2D grid map of the real scene in our POMCP modelling: black cells indicate visual occlusion, red cells indicate free agent positions and blue cells indicate candidate object positions.}
    \label{fig:intro_figure}
\end{figure}

This paper focuses on the Active Visual Search (AVS) problem in a known indoor environment.
We propose a motion planning policy that decides the movements of an agent within its observed world, in order to approach a specific object (the target) and visually detect it. 
\changemarker{When the target is successfully detected the agent can reach it following the shortest path} (see Figure~\ref{fig:intro_figure}(a)). 

AVS in real-world scenarios using an egocentric camera can be a very challenging problem due to the unpredictable quality of the observations, \ie object in the far field, motion blur and low resolution, partial views and occlusions due to scene clutters. This has an impact not only on the object detection but also on the planning policy.
To address this challenge, recent efforts are mostly based on deep Reinforcement Learning (RL), \eg deep recurrent Q-network (DRQN), fed with deep visual embedding~\cite{schmid2019iros,ye2019ral}. To train such DRQN models, a large amount of data is required, which are sequences of observations of various lengths, covering successful and unsuccessful search episodes from multiple real scenarios or simulated environments. 

Instead of performing any training to learn the policy beforehand, we propose to learn the AVS policy online to react properly at any environmental condition of the scene (\eg changes in the furniture), or to cope with the new modality of sensory data, without the need of an ad-hoc training.
This fundamental shift in the methodology is carried out considering the Partially Observable Monte Carlo Planning (POMCP) method~\cite{Silver2010}. POMCP has been applied in benchmark problems, such as rocksample, battleship and \textit{pocman} (partially observable pacman)  with impressive results, however, its use for robotic applications is an open and challenging research problem. To the best of our knowledge, this is the first attempt to use POMCP for the AVS problem. 

The overall architecture of \methname is shown in Figure~\ref{fig:method}.
At each time step, the inputs are the agent pose, \ie position and orientation, in a known 2D map and a RGB-D frame given by a sensor acquisition. An off-the-shelf object detector is applied to the RGB image, where the corresponding depth of the candidate target proposal is further exploited to obtain the candidate position in the map. Such object-related observation is then passed to the {\em POMCP exploration} module that assigns each possible move a reward indicating that a chosen move brings the agent closer to the object.
The policy is learnt online by Monte Carlo simulations and related particle-filter based belief update, therefore it is general and easy-to-deploy in any environment.
Crucially, our approach exploits the model of the environment to consider the sensor's field of view and all the admissible moves of the agent in the area. For our active vision search scenario, such a model can be easily obtained by building a map of the environment to include the position of fixed elements (\eg walls) but does not need to consider the position of moving objects.
Unlike other RL-based strategies \cite{schmid2019iros,ye2019ral}, implicitly encoding such environment knowledge in a data-driven manner, our motion policy explicitly use the knowledge of the environment for the visibility modelling.
Once the target is detected, the {\em robust visual approaching} module further localises the target on the map, so that a destination pose of the agent can be determined, \ie the closest pose with a frontal-facing viewpoint to the target, for the estimation of the shortest path~\cite{dijkstra1959note}. A path replanning scheme is proposed in the docking module to be robust to detector failures, such as miss-detections or false positives.

Our main contributions can be summarised as: 
1) we solve the policy learning bottleneck, which requires to have an offline training stage, with the first online policy learning by using the POMCP technique; 
2) we evaluate our approach according to Active Vision Dataset (AVD) benchmark \cite{ammirato2017dataset}, and show that it outperforms alternative approaches in terms of success rate without the cost of offline training under certain cases, and 
3) we perform an ablation study to assess the behaviour of the proposed approach when fed with increasingly corrupted detections and prove the robustness of our approach against missing detections.

\section{Related work}
\label{sec:soa}

Active Visual Search can be either addressed as a pure exploration task~\cite{Tovar2003,Ruangpayoongsak2005} \changemarker{where} target detection is merely subordinate to the solution of such task, or as an exploration and search task with target-specific inferences~\cite{Garvey1976,Wixson1994,Kollar2009,Sjoo2012,Aydemir2013,schmid2019iros,ye2019ral,han2019active}. One early \changemarker{example} of \changemarker{the latter approach} is indirect search~\cite{Garvey1976,Wixson1994,Kunze2014} which exploits intermediate objects (\eg a table) to restrict the search area for the target object (\eg a telephone). Although intermediate objects are usually easier to detect because of their size, their spatial relation w.r.t. the target may be not systematic. 
A softer reasoning is proposed in~\cite{Kunze2014}, where the likelihood of the target increases when objects which are expected to be co-occurring are detected. Such probabilistic modelling in a voxelised 3D scene representation is a common strategy to facilitate the planning of the agent's path towards the discovery of the target~\cite{Kollar2009,Shubina2010,Andreopoulos2011,Sjoo2012,Aydemir2013}, enriched by visual attention principles~\cite{Rasouli2017} that rank the search locations depending on their saliency.

AVS with deep learning is viable using Deep Reinforcement Learning techniques \cite{han2019active,schmid2019iros,ye2019ral}, where visual neural embeddings are often exploited for the action policy training. Han \etal{} \cite{han2019active} proposed a novel deep Q-network (DQN) where the agent state is given by CNN features describing the current RGB observation and the bounding box of the detected object. However, this work is limited since it assumes that the object has to be detected initially. To address the search task, EAT~\cite{schmid2019iros} performs feature extraction from the current RGB observation, and the candidate target crop generated by a region proposal network (RPN). The features are then fed into the Action Policy network. Twelve scenes from the AVD \cite{ammirato2017dataset} are used for the training of EAT.
Similarly, GAPLE~\cite{ye2019ral} uses deep visual features enriched by 3D information (from depth) for the policy learning.
Although GAPLE claims to be generalized, expensive training is the cost to pay as GAPLE is trained with 100 scenes rendered using a simulator House3D based on the synthetic SUNCG dataset. 
In general, RL-based strategies are dependent on the training with a large amount of data in order to encode the environmental modelling and motion policy. Differently, our proposed POMCP-based method makes explicit use of the available scene knowledge and performs efficient planning for the agent's path online without additional offline training.

As for optimal policy computation, a popular choice is to use 
\emph{Partially Observable Markov Decision Processes (POMDPs)}, a sound and complete framework for modeling dynamical processes in uncertain environments~\cite{Kaelbling1998}.
Computing exact solutions for non-trivial POMDPs is computationally intractable \cite{Papadimitriou1987}, but in the recent years impressive progress was made in developing approximate solvers. 
One of the most recent and efficient approximation methods for POMDP policies is \emph{Monte Carlo Tree Search (MCTS)}~\cite{Thrun2000,Coulom2006,Kocsis2006,Browne2012}, a heuristic search algorithm that represents system states as nodes of a tree, and actions/observations as edges. The most influential solver for POMDPs which takes advantage of MCTS is \emph{Partially Observable Monte Carlo Planning (POMCP)}~\cite{Silver2010} which combines a Monte Carlo update of the agent's belief state with an MCTS-based policy.
The most recent extensions of POMCP include applications to multiagents problems~\cite{Amato2015} and reward maximization while constraining the cost ~\cite{Lee2018}.
Finally,~\cite{Castellini2019a} uses the constraints on the state space to refine the belief space and increase policy performance. Here we build our AVS approach upon this method and propose a first methodology which integrates POMCP to AVS avoiding the training bottleneck of state-of-the-art AVS methods, allowing to move the agent and simultaneously learn the optimal policy.

\section{Method}
\label{sec:method}

We consider the scenario where an agent moves in a known environment, searching for a specific object. The agent explores the environment to find the target object, to localise it in the floor map, and then to approach it, \ie move close to the object location. 

The agent’s \emph{pose}\footnote{Here with pose we mean the 2D robot pose, \ie position and orientation, to be coherent with the related literature. We do not consider complex kinematics related to the agent structure (\eg if a humanoid robot is used).} at time step $t$ is $p_t=\{x_t,y_t,\theta_t\}$, where $x$ and $y$ are the coordinates on the floor plane, and $\theta$ is the \emph{orientation}.
At each time step the agent takes an action $a_t$:  It can move forward or backward, or rotate clockwise or counter-clockwise by a fixed angle. We assume the set of feasible actions is known a priori.
When the agent reaches a new pose $p_t$, it receives an observation which is the output of an object detector applied to the image acquired by a RGB-D camera\footnote{Notice that observations are not actions: \ie they are not actively performed by our POMCP planner, rather they are received after each movement of the robot.}.
We model the search space as a grid map (see Figure~\ref{fig:intro_figure}(b)).
Each cell can be either: 
(\emph{i}) ``\emph{visual occlusion}'', if the cell is occupied by obstacles, such as a wall or a piece of furniture, that prevent the agent to see through;
(\emph{ii}) ``\emph{empty}'', if the agent is allowed to enter the cell and thus no objects can be located in there; or 
(\emph{iii}) ``\emph{candidate}'', if none of the above, thus the cell is a candidate location for the target object.

We formulate the AVS problem as a Partially Observable Markov Decision Process (POMDP), which is a standard framework for modeling sequential decision processes under uncertainty in dynamical environments \cite{Kaelbling1998}. 
A POMDP is a tuple $(S,A,O,T,Z,R,\gamma)$, where $S$ is a finite set of partially observable \textit{states}, $A$ is a finite set of \textit{actions}, $Z$ is a finite set of \textit{observations}, $T$:~$S\times A \rightarrow \Pi(S)$ is the \textit{state-transition model}, $O$:~$S\times A \rightarrow \Pi(Z)$ is the \textit{observation model},  $R$:~$S \times A \rightarrow \R$ is the \textit{reward function} and $\gamma \in [0,1)$ is a \textit{discount factor}.
Agents operating POMDPs aim to maximise their expected total discounted reward 
$E[\sum_{t=0}^{\infty} \gamma^t R(s_t,a_t)]$, by choosing the best action $a_t$ in each state $s_t$, where $t$ is the time instant; $\gamma$ reduces the weight of distant rewards and ensures the (infinite) sum's convergence. The partial observability of the state is dealt with by considering at each time-step a probability distribution over all the states, called the \emph{belief} $B$. 
POMDP \emph{solvers} are algorithms that compute, in an exact or approximate way, a \emph{policy} for POMDPs, namely a function $\pi$:~$B \rightarrow A$ that maps beliefs to actions. 

\begin{figure*}[t!]
  \centering
  \includegraphics[width=.95\textwidth]{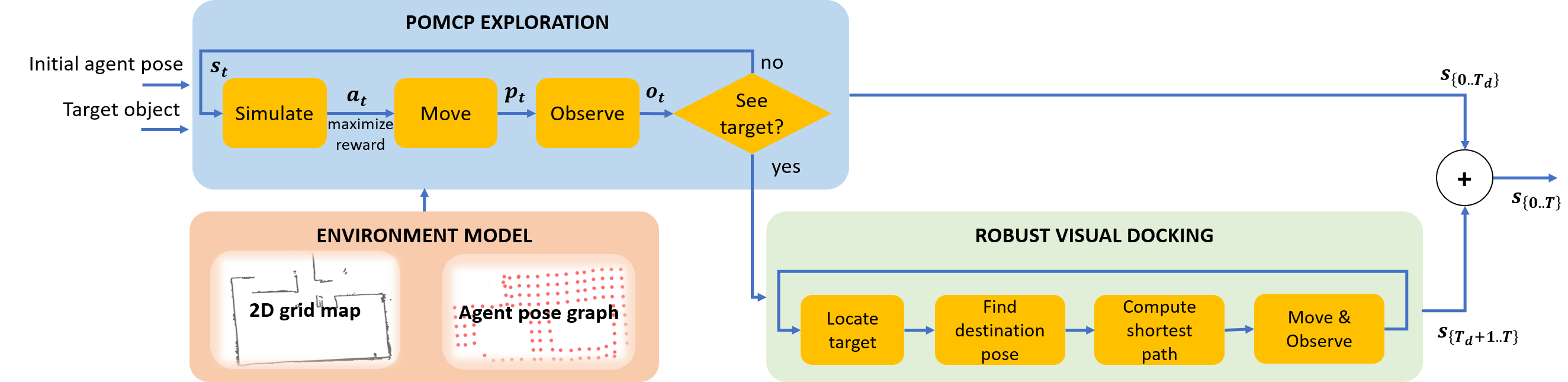}
  \caption{Overall architecture of our proposed method \methname. The red box represents prior knowledge pushed into the POMCP module, the blue box represents the exploration strategy to detect the target object, and the green box represents the visual docking strategy to reach the destination pose. Math notation: state $s_t$, action $a_t$, pose $p_t$, observation $o_t$, POMCP state sequence $s_{\{0..T_d\}}$, docking state sequence $s_{\{T_{d+1}..T\}}$, complete state sequence $s_{\{ 0..T\}}$.} 
  \label{fig:method}
  \vspace{-0.25cm}
\end{figure*}

We therefore propose \methname to address the POMDP problem with a Monte-Carlo Tree Search strategy that allows us to learn the policy online (POMCP). A graphical overview of the method is shown in Figure~\ref{fig:method}. The POMCP exploration module explores the known environment to detect the target with some prior knowledge that can be obtained from the pre-exploration of the environment. The learning process ends when the agent detects the object. 
Once the target object is detected, we can approach it. We first localise the detected target using the depth channel together with the camera pose (which can be obtained by the agent pose with agent-camera calibration). With the target position, we then compute the destination pose as the closest pose of the agent to the object, facing it frontally.
Finally, we drive the agent to reach the target pose by using a shortest path exploration method with a path replanning scheme to be robust against imperfect detectors.
We will detail our framework in the following sections.

\subsection{POMCP exploration} \label{sec:pomdp}
%
\emph{Partially Observable Monte Carlo Planning (POMCP)}~\cite{Silver2010} is an online Monte-Carlo based solver for POMDPs. It uses \emph{Monte-Carlo Tree Search} (MCTS) for selecting, at each time-step, an action which approximates the optimal one.
The Monte Carlo tree is generated by performing a certain number of simulations ($nSim$) from the current belief. 
A big advantage of POMCP is that it enables to scale to large state spaces because it never represents the complete policy but it generates only the part of the policy related to the belief states actually seen during the plan execution. Moreover, the local policy approximation is generated online using a simulator of the environment, namely a function that given the current state and an action provides the new state and an observation according to the POMDP transition and observation models.

The methodology proposed here is a specialization of POMCP for the AVS problem. It is based on four main elements defined in the following and used all together by POMCP to perform the search of an object in the environment. We assume that $n$ is the number of possible poses -- \ie pairs (positions, orientations) -- that the agent can take in the environment,
$m$ is the number of objects in the environment, and $k$ is the number of positions in which each object can be located.
%
The first element of the proposed framework is a pose graph $\mathcal{G}$ in which nodes represent the $n$ possible poses of the agent and edges connect only poses reachable by the agent with a single action.
The second element is the set $\mathcal{H} = \{ 1, \ldots, k \}$ of all possible indices of positions that each objects can take in the environment.
Each index in $\mathcal{H}$ corresponds to a specific position in the topology of the environment where the search is made. The third element of our framework is the hidden state of the system, which is represented by a vector of object positions $\mathcal{P} = \{ p_1, \ldots, p_m \}$, where $p_i \in \mathcal{H}$ indicates the pose of the $i$-th object in the environment. The goal of the search is to reach a specific object. 
The fourth element is a matrix of object visibility $\mathbf{L} = (l_{i,j}) \in \{0,1\}^{n \times m}$, where $l_{i,j}=1$ if the object $j$ is visible from pose (\ie agent's position and orientation) $i$. Matrix $\mathbf{L}$ can be deterministically derived from $\mathcal{G}$, $\mathcal{H}$ and $\mathcal{P}$ by a visibility function $f_L$ which computes the visibility of each object from each agent pose, considering the physical properties of the environment. 

POMCP uses all these elements during its computation: Vectors of object poses $\mathcal{P}$ are first used to represent possible hidden states (\ie possible arrangements of objects in the environment), these vectors are then used to generate matrices $\mathbf{L}$ of object visibility that are used, together with graph $\mathcal{G}$, to perform simulation steps. In particular, at each step the POMCP simulator takes the current agent pose $\bar{i}$ (\ie node of $\mathcal{G}$) and computes the related set of visible objects $\{ j \ | \ l_{\bar{i},j} = 1, l_{\bar{i},j} \in \mathbf{L} \}$. 
If this set of objects contains the searched object than a positive reward is provided, the search involving POMCP is {\em terminated}, 
otherwise a negative reward is provided (corresponding to the energy spent to perform the movement) and the POMCP-based search is continued. To prevent the agent to visit the same poses more than once, the agent maintains an internal memory vector that collects all the poses already visited during the current run. Every time the agent visits a pose already visited it receives a high negative reward.
The planner gets an observed value $1$ if the searched object has been observed, $0$ otherwise. The belief of the agent at each step is an (approximated) probability distribution over positions of the searched object in the environment, that represents the POMCP hidden state.
\changemarker{If after a given amount of moves, the object is not observed, the method terminates and reports a search failure.}
\subsection{Robust Visual Docking}
\label{sec:visualdocking}

Once the agent has explored the environment and detects the target object, the agent is asked to approach the object and stop as close as possible in front of it; this pose is named \emph{destination pose}.
We first process the depth channel of the last observation to estimate the position of the detected target in the environment. The depth crop of the object detection is converted to 3D points with the camera pose, where the $\lbrace x, y\rbrace$ position of the closest point to the camera is used to approximate the target position.
Then we generate the destination pose by selecting the subset of admissible poses that can see the target position, according to the observation model, and finding the closest to the target position.
We use the Dijkstra algorithm~\cite{dijkstra1959note} to compute the shortest path between the current pose (reached using POMCP) and the estimated destination pose.
In order to be robust against the detector imperfections, we further introduce a path replanning scheme after new observations. At every time step in the approaching phase the agent observes the scene. In case the target object is detected, we recompute the object location in the environment, the destination pose and re-plan the optimal path to reach it. If instead the object is not detected, the agent continues with the planned path.
The effect of the path replanning against following the originally planned path can be seen in Table~\ref{tab:res_comparison_GT_merged}, where we have an average improvement of 10\% in success rate.

\section{Experiments}
\label{sec:exp}

\begin{figure*}[!t]
\begin{center}
	\begin{tabular}{@{}c@{}c@{}c}
		\includegraphics[width=0.3\textwidth]{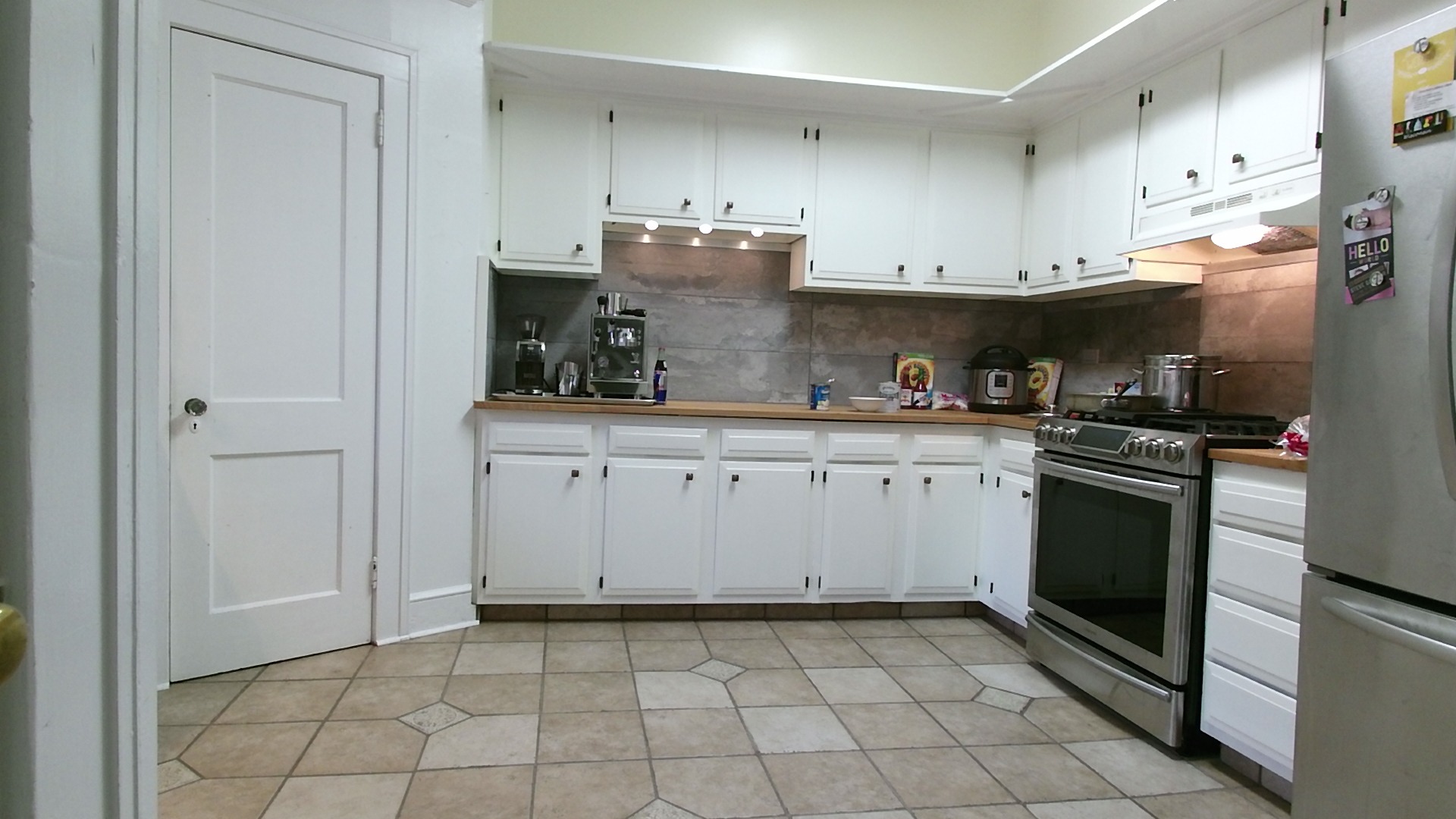}&
		\includegraphics[width=0.3\textwidth]{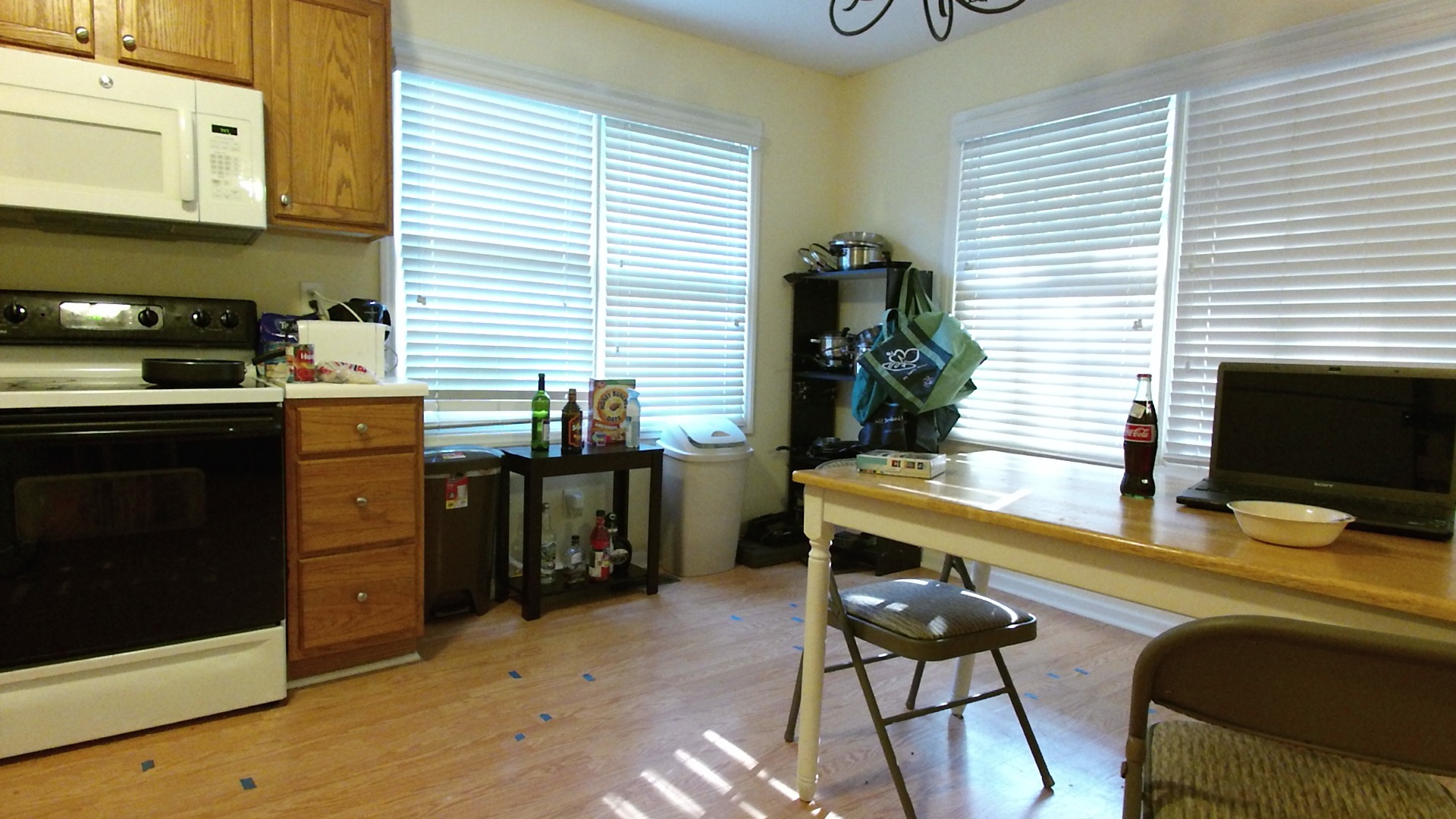}&
		\includegraphics[width=0.3\textwidth]{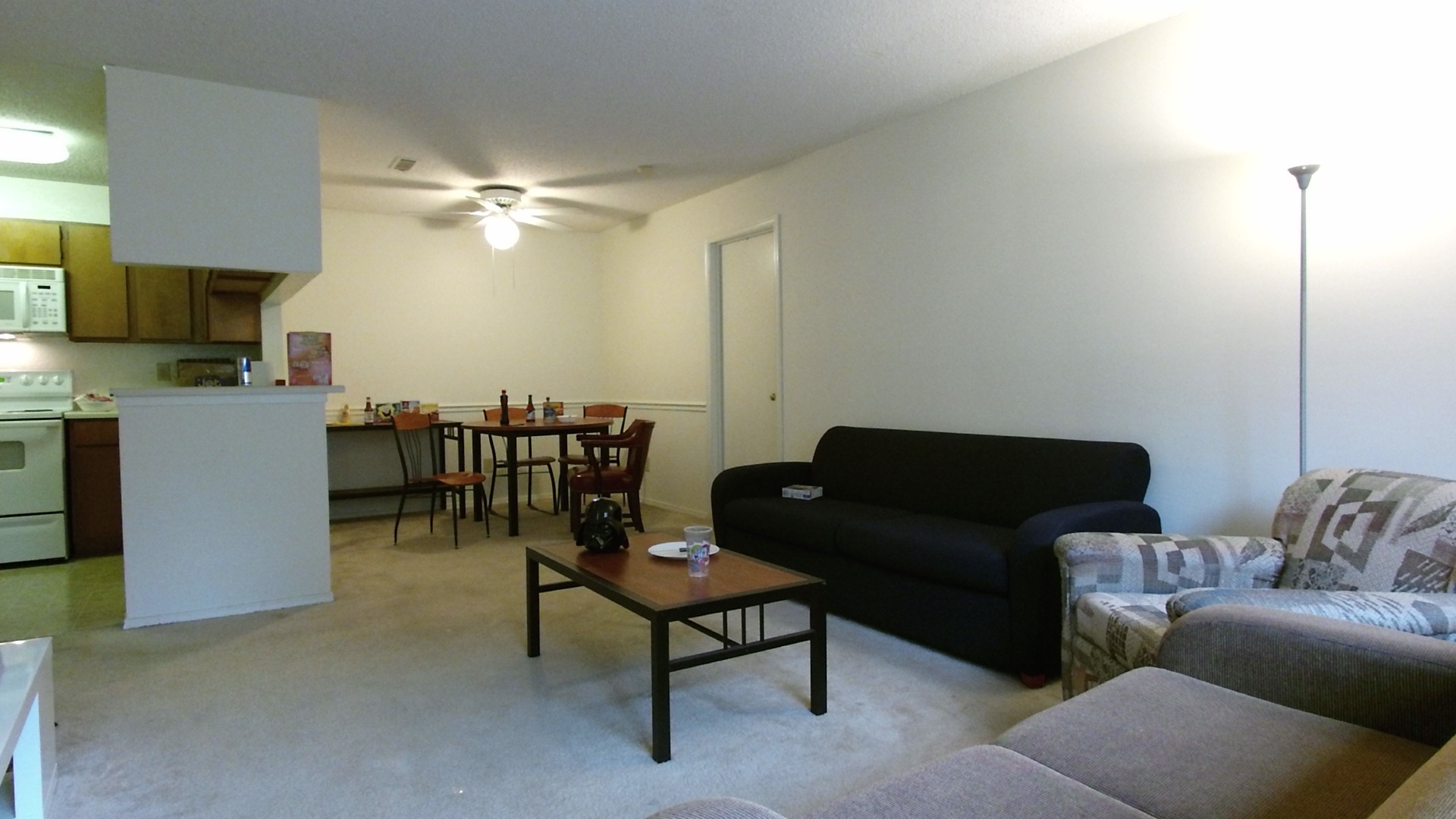}\\
		(a) Easy: Home\_005\_2 & (b) Medium difficult: Home\_001\_2 & (c) Hard: Home\_003\_2 
   	\end{tabular}
\end{center}
\vspace{-0.3cm}
\caption{Test scenes from AVBD of 3 different difficulty levels (as in \cite{schmid2019iros}). (a) the agent explores within a kitchen, (b)  the agent explores in one living room with an open kitchen, and (c) the agent explores one large living room with a half-open kitchen and dinning area.}
\label{fig:exp_scenes}
\end{figure*}

We validate our proposed method against baselines and state-of-the-art methods using the AVD dataset~\cite{ammirato2017dataset} following the evaluation protocol defined by the AVD Benchmark (AVDB) on the task of active object search in \emph{known environments} (referred as \emph{Task 1a} in the benchmark). AVD is the largest real-world dataset available for testing active visual search, containing scans of 14 real apartments recorded using a robot equipped with a RGB-D camera, so allowing for a virtual exploration of the environment. 
\vspace{-1em}
\paragraph*{Test scenes:}Following the analysis proposed by the EAT authors~\cite{schmid2019iros}, we test three scenes that correspond to three different difficulty levels (see Figure~\ref{fig:exp_scenes}). The easy level is represented by Home\_005\_2, where the agent only explores within a kitchen area. The medium difficult scene is represented by Home\_001\_2, where the agent explores in a living room with an open kitchen. Finally Home\_003\_2 represents the most difficult scene where the agent explores a large living room with a half-open kitchen and some dining area. 
For each scene, the agent's pose graph and the ground-truth (GT) annotations of each target object are provided by AVD, while we prepare the 2D grid map of each scene for the POMCP module. To obtain the occluded cells, we first perform 3D scene reconstruction using Open3D~\cite{Zhou2018} followed by a z-plane intersection.



\vspace{-1em}
\paragraph*{Comparison:} We remark that with \textbf{\methname} 
we are introducing a new (harder) scenario where no training is allowed, so no other published approaches are available as comparison.
Nevertheless we refer to state-of-the-art systems which applied on the AVDB, quoting their performances in italic, to remind that they derive from an easier setup.  
In particular, we consider the RL-based EAT~\cite{schmid2019iros} and GAPLE~\cite{ye2019ral} (discussed in  Sec.~\ref{sec:soa}). Unfortunately, the protocol adopted with GAPLE on AVD is not documented, while in~\cite{schmid2019iros} their protocol is explicit, but is different from that of the AVD benchmark. With EAT, only a subset of objects are used for the searching task in each scene, while the AVDB protocol uses all objects. To this sake, we mark with an asterisk ${}^{(*)}$ the numbers obtained with the EAT protocol (reported in the original paper~\cite{schmid2019iros}).
As comparative baselines we consider two methods: the \textbf{Random Walk}, that allows the agent to randomly select an action among all the feasible ones at each time step.  The second baseline is \textbf{partial-\methname}, \ie an ablation of \methname, where we exclude the path replanning after new observations. This helps to appreciate the net contribution of the path replanning scheme during the visual docking phase.

\vspace{-1em}
\paragraph*{Evaluation metrics:} In line with the AVDB, we consider: \emph{Success Rate (SR)} \ie the percentage of times the agent successfully reaches one of the destination poses (as provided in AVDB) over the total number of trails (a larger value indicates a more effective search);  
\emph{Average Path Length (APL)} is the average number of poses visited by the agent among the paths that lead to a successful search (a lower value indicates a higher efficiency); Finally, the \emph{Average shortest path length (ASPPL)} is the average ratio between the shortest possible path to reach a valid destination pose (provided by AVDB as a piece of GT information) and the length of the path generated by the model (a larger value indicates a higher absolute efficiency). 
Additionally, we compute the standard deviation of ASPPL to investigate the variability of \methname in behaving efficiently. 

\vspace{-1em}
\paragraph*{Result Discussion:} 
\begin{table}[t]
\caption{Results on the three test scenes from AVDB using the object detections from GT annotations. EAT numbers are in italic to remind its easier setup (training is permitted). The asterisks ${}^{(*)}$ indicate the EAT protocol with less objects into play (as published in~\cite{schmid2019iros}) and this gives the first plate of the table. The bottom plate reports numbers obtained with the full original AVDB protocol (more objects into play). The parentheses show standard deviations.}
\label{tab:res_comparison_GT_merged}
\resizebox{\textwidth}{!}{%
\begin{tabular}{|c|c|c|c|c|c|c|c|c|c|c|c|c|}
\hline
\multirow{2}{*}{} & \multicolumn{3}{c|}{Easy} & \multicolumn{3}{c|}{Medium} & \multicolumn{3}{c|}{Hard} & \multicolumn{3}{c|}{Avg.} \\ \cline{2-13} 
 & SR & APL & ASPPL & SR & APL & ASPPL & SR & APL & ASPPL & SR & APL & ASPPL \\ \hline
{\em EAT}~\cite{schmid2019iros}$^{(*)}$  & \emph{0.77} & \emph{12.2} & - & \emph{0.73} & \emph{16.2} & - & \emph{0.58} & \emph{22.1} & - & \emph{0.69} & \emph{16.8} & - \\ \hline
Random Walk$^{(*)}$ & 0.32 & 74 & 0.19 (0.35) & 0.113 & 74.48  & 0.21 (0.36)  & 0.10 & 79.27 & 0.17 (0.18) & 0.18 & 75.91 & 0.19 (0.29)  \\ \hline
partial-\methname$^{(*)}$ & 0.96 & 12.88 & 0.73 (0.25) & 0.68 & 16.13 & 0.80 (0.23)  & 0.41 & 21.05 & 0.70 (0.40) & 0.68 & 16.68 & 0.74 (0.29)\\ \hline
\methname$^{(*)}$ & 0.98 & 13.6 & 0.72(0.26) & 0.73 & 17.1 & 0.8(0.23) & 0.56 & 20.5 & 0.72(0.39) & 0.76 & 17.1 & 0.75 (0.29) \\ 
\hline
\hline
Random Walk & 0.22 & 71.47 & 0.23(0.38) & 0.16 & 69.84  & 0.22 (0.33)& 0.14 & 62.30 & 0.29(0.38)  & 0.17  & 67.87 & 0.25 (0.36) \\  \hline
partial-\methname & 0.88 & 12.19 & 0.80 (0.24) & 0.68 & 16.75  & 0.74(0.24) & 0.34 & 23.09 & 0.66 (0.33)& 0.63 & 17.34 & 0.73 (0.27) \\ \hline
\methname & 0.93 & 12.96 & 0.78 (0.24) & 0.80 & 18.2 & 0.72(0.24) & 0.43 & 21.9 & 0.65 (0.32)& 0.72 & 17.68 & 0.72 (0.26) \\ \hline
\end{tabular}%
}
\end{table}
\begin{table}[t]
\caption{Results on the three test scenes using the object detector provided by AVDB, and its original protocol (all the objects into play) \cite{ammirato2018target}.}
  \centering
     \resizebox{\textwidth}{!}{%
    \begin{tabular}{|c|c|c|c|c|c|c|c|c|c|c|c|c|}
    \hline
    \multirow{2}{*}{} & \multicolumn{3}{c|}{Easy} & \multicolumn{3}{c|}{Medium} & \multicolumn{3}{c|}{Hard} & \multicolumn{3}{c|}{Avg.} \\ \cline{2-13} 
     & SR & APL & ASPPL & SR & APL & ASPPL & SR & APL & ASPPL & SR & APL & ASPPL \\ \hline
    Random Walk & 0.22 & 71.47 & 0.23(0.38) & 0.16 & 69.84  & 0.22 (0.33)& 0.14 & 62.30 & 0.29(0.38)  & 0.17  & 67.87 & 0.25 (0.36)\\  \hline
     partial-\methname & 0.61  & 17.49 & 0.7 (0.29) & 0.37 & 19.2 & 0.64(0.26) & 0.18 & 26.22 & 0.54(0.28) & 0.38 & 20.97 & 0.62 (0.27) \\ \hline
    \methname &  0.6 & 17.9 & 0.68 (0.28) & 0.40 &  20.73 & 0.62 (0.26) & 0.19 &  26.6 & 0.53 (0.28) & 0.4 & 21.74 & 0.61(0.27) \\ \hline
    \end{tabular}%
}
    \label{tab:res_detector}
\end{table}
Table~\ref{tab:res_comparison_GT_merged} is divided in two plates, the upper showing the results obtained with the EAT~\cite{schmid2019iros} protocol (less objects into play, approaches marked with an asterisk ${}^{(*)}$), and where the EAT numbers are in italic to remind that EAT has been trained beforehand on separate data, while we are training-free. The plate below shows numbers obtained with the original AVDB protocol (all the objects are considered). 
On the right we have the average of the performance of the three scenarios (mean of the means) and the average of the three standard deviations. To compare the exploration engine of \methname, discarding nuisances caused by the underlying object detectors, we report the results using an ideal detector (as in EAT~\cite{schmid2019iros}). The behavior of \methname when in presence of noisy detectors is the subject of another experiment.  

From Table~\ref{tab:res_comparison_GT_merged} we can see that on average our proposed method is able to outperform EAT in terms of the SR with a comparable APL, despite our setup eliminates any training. Notably, our proposed \methname has a dominant advantage against EAT in terms of SR, with this advantage decreasing as the scene gets more difficult.  



\begin{figure}[t!]
\begin{center}
	\begin{tabular}{@{}c@{}c}
		\includegraphics[width=0.45\textwidth]{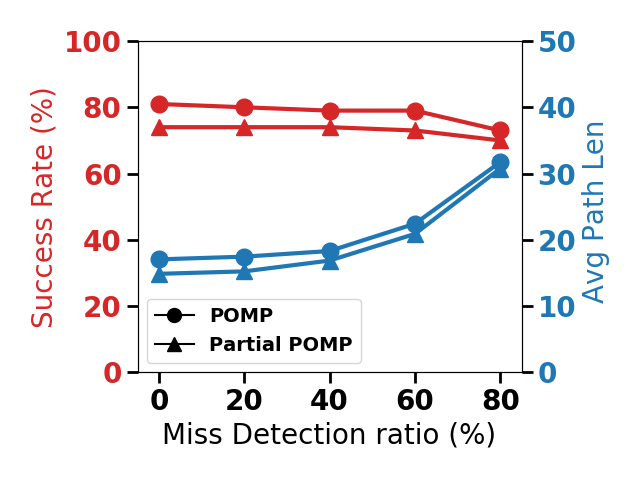}&
		\includegraphics[width=0.45\textwidth]{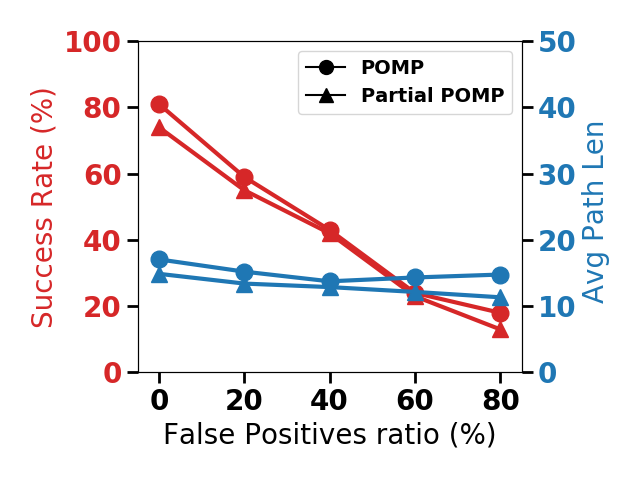}
   	\end{tabular}
\end{center}
\vspace{-0.5cm}
\caption{Results of our proposed methods using the medium difficult scene (Home\_001\_2) over various ratios of missing detections and false positives. \changemarker{Shown is the Success Rate (in Red) and Average Path Length (in Blue)}} 
\label{fig:detector_study_GT}
\end{figure}

Table \ref{tab:res_detector} shows the performance of \methname against the baselines with a detector provided by the same authors of the AVDB \cite{ammirato2018target}. 
The detector is similar to Faster-RCNN\cite{ren2015faster} but with additional input as the reference images of the target object, in order to improve the detection quality.
We use the pre-trained model without any customisation. The detector achieves a precision of 0.73 and a recall of 0.53 under the confidence threshold of 0.9. On average, our method with path replanning improves the SR with a slight increase in APL, which is consistent with what we observe in Fig.~\ref{fig:detector_study_GT}. 
In terms of processing speed, we run experiments on a machine with 6-core Intel i7-6800k CPU, achieving 0.07 seconds per step on average.

Since the detector plays a role in \methname, we investigate its impact in terms of missing detections and false positives by manipulating the GT annotations. Specifically, for each target in a scene, we randomly exclude a set of ratios, from 0\% to 80\% with 20\% as a gap, of its GT annotations. Regarding the false positives, we randomly change the label of detections corresponding to other instances to the target object, for the set of ratios from 0\% to 80\% with 20\% as a gap, of its GT annotations. Fig.~\ref{fig:detector_study_GT} shows the plot of \methname and partial-\methname in terms of SR and APL over a set of varying ratios of missing detections (left) and false positives (right). The results are averaged over 551 independent runs composed by 19 target objects and 29 starting positions for each target object. On one hand, we see that both the versions are robust to missing detections where the SR only starts to noticeably decrease after 60\% missing detections. The path length starts to have a noticeable increase after 40\% missing detections. On the other hand, we are vulnerable to false positives in terms of the SR, while the APL is not much affected. This is because more false positives lead to a higher chance of POMCP perceiving wrong destinations and ending up with failure paths. However, since the APL is only computed among successful paths, the impact of false positive ratio is therefore limited. From both plots, we can also prove that \methname with path replanning in the shortest path computation can further boost the SR although with a trade-off of an increase in APL. 

\section{Conclusions}
\label{sec:conclusion}

We proposed a POMCP-based planner, POMP, to learn an optimal policy for AVS in known indoor environments. To the best of our knowledge, our approach is the first to use an online policy learning method for AVS. Notably, POMP does not need expensive (in time and computation) labelled data but rather exploits the information of the floormap of the environment, which is usually available or easy to obtain (\eg via a single exploration run). We evaluated our approach following the AVD benchmark and achieved comparable performance (\ie average success rate and average path length) against the state-of-the-art methods while using far less information.
This work paves the way to several interesting research directions, including the possibility of integrating more scene priors, \eg object co-occurrence, in the POMCP modelling to further boost the performance.




\bibliography{refs}

\end{document}